\documentclass{article}

    \PassOptionsToPackage{numbers, compress}{natbib}

\usepackage[preprint]{neurips_2026}

\usepackage[utf8]{inputenc} 
\usepackage[T1]{fontenc}    
\usepackage{hyperref}       
\usepackage{url}            
\usepackage{booktabs}       
\usepackage{amsfonts}       
\usepackage{nicefrac}       
\usepackage{microtype}      
\usepackage{xcolor}         

\usepackage[utf8]{inputenc}
\usepackage{lmodern}
\usepackage{amsmath,amssymb,amsthm,mathtools}
\usepackage{bm}
\usepackage{multirow}
\usepackage{algorithm}
\usepackage{algpseudocode}
\usepackage{tabularx} 

\usepackage[most]{tcolorbox}

\usepackage{caption}
\usepackage{placeins}

\makeatletter
\newtcolorbox{breakablealgorithm}[2][]{%
  enhanced,
  breakable,
  frame hidden,
  colback=white,
  boxsep=0pt,
  left=0pt,
  right=0pt,
  top=2pt,
  bottom=2pt,
  before upper={%
    \refstepcounter{algorithm}%
    \textbf{Algorithm~\thealgorithm}\ #2\par
    \vspace{2pt}\hrule\vspace{4pt}%
  },
  overlay unbroken={%
    \draw[line width=0.8pt] (frame.north west) -- (frame.north east);
    \draw[line width=0.8pt] (frame.south west) -- (frame.south east);
  },
  overlay first={%
    \draw[line width=0.8pt] (frame.north west) -- (frame.north east);
    \draw[line width=0.8pt] (frame.south west) -- (frame.south east);
  },
  overlay middle={%
    \draw[line width=0.8pt] (frame.north west) -- (frame.north east);
    \draw[line width=0.8pt] (frame.south west) -- (frame.south east);
  },
  overlay last={%
    \draw[line width=0.8pt] (frame.north west) -- (frame.north east);
    \draw[line width=0.8pt] (frame.south west) -- (frame.south east);
  },
  #1
}
\makeatother

\title{Stable Long-Horizon PDE Forecasting via Latent Structured Spectral Propagators}

\author{
  Xiaoxiao Lu \qquad Ye Yuan\thanks{Corresponding author.} \qquad Jiahao Shi\\
   School of AIA, Huazhong University of Science and Technology\\
  \texttt{\{xiaolu, yye, jiahaoshi\}@hust.edu.cn}
}

\begin{document}

\maketitle

\begin{abstract}
Long-horizon forecasting of time-dependent partial differential equations (PDEs) is critical for characterizing the sustained evolution of physical systems. While neural operators have emerged as efficient surrogates, they typically learn implicit finite-time transitions from discrete observations. When deployed autoregressively, such propagators often suffer from rapid error accumulation and dynamic drift. To address this, we propose a neural forecasting framework that reformulates PDE rollout as learning a Structured Spectral Propagator (SSP) in a propagation-oriented latent space. Following an analysis-propagation-synthesis design, our framework: (i) maps physical states into a shared, time-consistent spatial representation; (ii) projects this space into a compact propagation state to isolate recurrent dynamics from fine-grained spatial details, thereby decoupling reconstruction fidelity from rollout regularity; and (iii) evolves retained spectral modes using a frequency-conditioned linear backbone complemented by a nonlinear spectral closure to account for truncated interactions. This explicit structuring endows the propagator with a strong inductive bias for coherent modal evolution. Extensive experiments demonstrate that SSP significantly outperforms state-of-the-art baselines, reducing relative $L_2$ errors by up to 48.9\% and exhibiting improved stability in temporal extrapolation beyond the supervised horizon. 
\end{abstract}

\section{Introduction}

Modeling physical dynamics governed by time-dependent partial differential equations (PDEs) is a central problem in computational science. Data-driven neural surrogates, particularly neural operators, have emerged as efficient alternatives for approximating PDE evolution~\cite{deep_learning_for_physical,ml_pde,deeponet,neural_operator_zongyi,FNO,CNO,GNOT}.  In practice, these models are commonly trained from discrete observations to learn finite-time propagation maps and then deployed autoregressively~\cite{stencil_modeling,MP-PDE,TL-DEEPonet,RNO}. However, many scientific applications require long-horizon forecasts to characterize the long-time behavior of physical systems. Such forecasts are obtained by iteratively composing the learned propagator, with each predicted state fed back as input. In this recursive regime, even tiny one-step approximation errors can be nonlinearly amplified, causing the model to drift away from the manifold of physical solutions~\cite{MP-PDE,TL-DEEPonet,RNO,difficulty_roll_chaotic}. Therefore, the goal is not merely to fit finite-time transitions but to learn a propagator whose internal update remains structured and reliable under repeated composition.

Existing neural PDE surrogates approach long-horizon prediction from several directions. Rollout-oriented methods~\cite{stencil_modeling,MP-PDE, RNO,pde_refiner,wu2025coast} mitigate error accumulation through recurrent supervision, temporal stencils, or refinement, while the recurrent transition itself often remains an implicit map with limited structure. Latent-space methods~\cite{LGE,LNS,LNO,AROMA,calm,UPT} reduce evolution complexity by learning compact representations, and Koopman-inspired models~\cite{KNO,IKNO,koopman_pde_ref} introduce linear dynamics priors in learned observable spaces. However, weak geometric constraints on the latent space create an intrinsic conflict between representation richness, dynamical simplification, and stability. Accurate reconstruction often demands high-dimensional latent spaces to preserve fine-grained spatial details, yet such complexity frequently compromises the regularity and identifiable structure of the temporal advancement.  

A natural question is how to endow structural constraints on the propagator for stable recursive application. We take the spectral properties of PDE evolution as an inductive bias for propagation law learning~\cite{dft_pde_linear}. For translation-invariant linear PDEs, temporal advancement decomposes into independent low-dimensional linear systems in the spatial Fourier basis, suggesting that propagation may admit a more explicit internal organization in a suitable representation. For general nonlinear PDEs, such regularity is no longer available in the original domain, motivating us to learn a time-consistent spatial representation in which the dynamics admit a more tractable description. 

We propose Structured Spectral Propagator (SSP) for long-horizon PDE forecasting in propagation-oriented latent spaces. SSP encodes each time slice into a shared, time-consistent spatial-basis representation and projects it into a compact propagation state. This compact state forms the propagation-oriented space, where redundant degrees of freedom from the rich spatial representation are suppressed. The propagator advances this state by evolving a retained set of latent Fourier modes with a frequency-conditioned linear backbone, complemented by a nonlinear spectral closure that accounts for residual cross-mode interactions and truncation-induced effects. The inverse projection lifts the propagated state back to the spatial-basis space, and the decoder synthesizes the physical field. This design separates spatial representation construction from temporal advancement, yielding a compact and structured update law for recursive rollout. 

Our contributions are summarized as follows: (i) we formulate long-horizon PDE forecasting as structured propagator learning under repeated autoregressive composition; (ii) we separate spatial-basis representation from temporal propagation through a frame-wise encoder and a compact propagation state, reducing the conflict between reconstruction fidelity and rollout regularity; (iii) we instantiate structured spectral propagation with a frequency-conditioned linear modal backbone and nonlinear spectral closure, and validate the design on long-horizon and beyond-horizon PDE forecasting benchmarks.


\section{Background}

\paragraph{Partial differential equations.}
Let \(\Omega\subset\mathbb R^d\) be a spatial domain with boundary \(\partial\Omega\), and let \(u:\Omega\times[0,T_{\max}]\to\mathbb R^{d_u}\) denote the solution field. We consider the evolution of physical fields governed by time-dependent PDEs:
\begin{equation}
F\left(x,t,\left\{\partial_t^{r}D_x^{\alpha}u(x,t)\right\}_{0\le r\le s_t,\ |\alpha|\le s_x}\right)=0,
\qquad (x,t)\in\Omega\times(0,T_{\max}],
\label{eq:general_pde}
\end{equation}
where \(D_x^\alpha\) denotes a spatial derivative with multi-index \(\alpha\), and \(s_t\) and \(s_x\) denote the maximal temporal and spatial derivative orders. The operator \(F\) may contain both linear and nonlinear differential terms. The equation is equipped with initial conditions \(\partial_t^j u(x,0)=u_j(x)\) for \(j=0,\ldots,s_t-1\), and boundary conditions \(\mathcal B[u](x,t)=g(x,t)\) on \(\partial\Omega\times(0,T_{\max}]\).

\paragraph{Neural operators as finite-time propagators.}
Let \(\mathcal U\) and \(\mathcal V\) be Banach spaces of input and solution functions. Operator learning~\cite{FNO,U-FNO,UNO,F-FNO,jin2022mionet,pino,GINO} aims to approximate a ground-truth operator \(\mathcal G:\mathcal U\to\mathcal V\) by a neural operator \(\widehat{\mathcal G}:\mathcal U\to\mathcal V\). For time-dependent PDEs, this operator is typically learned through finite-time prediction between solution snapshots or short temporal windows~\cite{KNO}. Long-horizon forecasting then requires recursively composing the learned propagator to approximate the global evolution induced by \(\mathcal G\). Therefore, the learning problem is not fully characterized by fitting individual finite-time transitions: the learned map must also remain meaningful and reliable when repeatedly applied during rollout. This motivates learning a time-consistent representation in which the evolving state admits a shared, structurally constrained propagation law across rollout steps.

\paragraph{Latent-space operator factorization.}
Latent-space methods~\cite{LGE,LNS,LNO,AROMA,calm,UPT,chen2022crom,oplnf,PI-ROM,mamba} approximate the operator through a factorized map, 
\(\mathcal{G} \approx \widehat{\mathcal G}:=\mathcal D\circ\Phi\circ\mathcal E\), where the encoder \(\mathcal E:\mathcal U\to\mathcal Z\) maps input functions to a latent space, the processor \(\Phi:\mathcal Z\to\mathcal Z\) evolves latent states, and the decoder \(\mathcal D:\mathcal Z\to\mathcal V\) maps them back to the solution space. Such encoder--processor--decoder designs are commonly used to reduce computational cost, enable scalable latent processing, and improve flexibility across discretizations or geometries. Existing methods such as LNO~\cite{LNO}, CALM-PDE~\cite{calm}, AROMA~\cite{AROMA}, and hybrids based on proper orthogonal decomposition (POD) or reduced-order modeling (ROM)~\cite{chen2022crom,POD_2,POD_3} exploit compact latent coordinates for efficiency and adaptation. Despite their effectiveness, compact latent spaces are often designed primarily as computational representations, rather than explicitly structured to support time-consistent recursive propagation.


\paragraph{Koopman-inspired propagation constraints.}
Koopman-inspired methods~\cite{KNO,IKNO,koopman_pde_ref} further constrain latent evolution by parameterizing \(\Phi\) to advance learned observables in an approximately linear manner. This follows the Koopman view that a nonlinear flow induces a linear operator on a suitable space of observables. Nevertheless, the learned latent space is often used simultaneously for reconstruction and propagation, creating a tension between spatial richness and dynamical regularity. Moreover, when temporal context within an input window is used to construct the latent representation~\cite{KNO}, predictive supervision may allow the encoder to encode short-term advancement into the representation on which \(\Phi\) operates, making the latent update less identifiable as a standalone propagation rule across rollout steps. This motivates a cleaner separation between time-consistent spatial representation and compact, structured propagation.



\begin{figure}[t]
		\centering
		\includegraphics[width=.99\linewidth]{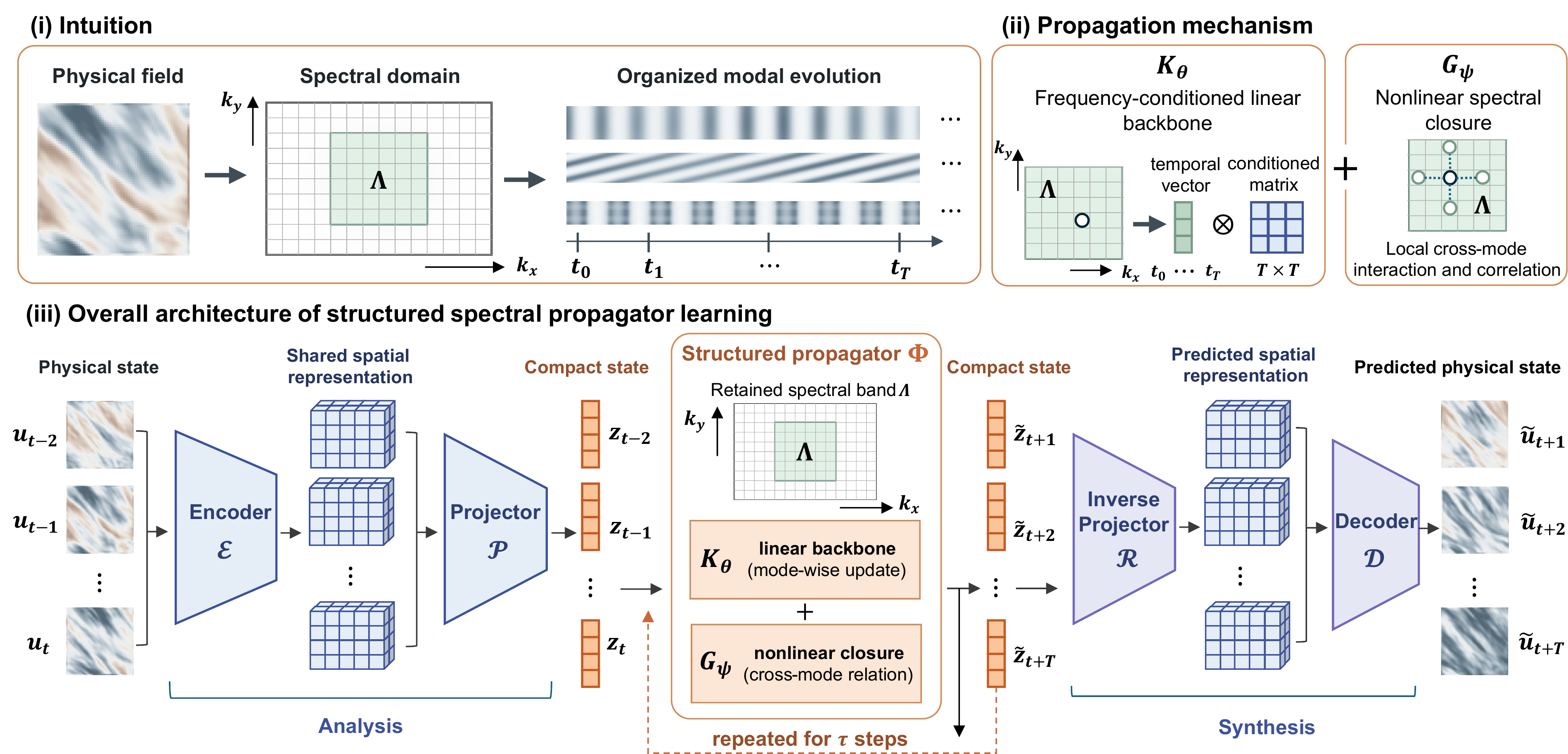}
        \caption{
        Overview of the proposed SSP framework. (i) Spectral intuition: a suitable spatial basis exposes organized modal evolution in the spectral domain. (ii) Propagation mechanism: the structured propagator combines a frequency-conditioned linear backbone \(K_\theta\) for mode-wise temporal updates with a nonlinear closure \(G_\psi\) for local cross-mode interactions within the retained spectral band. (iii) Overall architecture: SSP follows an analysis--propagation--synthesis pipeline, encoding physical states into a shared spatial representation, projecting them to compact propagation states, performing structured spectral propagation, and decoding the lifted representations into future physical fields.
        }
		\label{spectral-decouple}
\end{figure}

\section{Architecture Desiderata}

The preceding discussion shows that stable long-horizon forecasting requires more than fitting local transitions. We use the spectral structure of PDE evolution as a guiding principle for designing an explicitly organized recurrent propagator.


\noindent\textbf{Spectral decomposition of the linear propagation operator.}
Consider the linear evolution equation \(\partial_t u(x,t)=\mathcal L_x u(x,t)\) on the periodic domain \(\Omega=\mathbb T^2\), where \(\mathcal L_x\) is linear and translation-invariant in space. Under the spatial Fourier basis, the state decomposes into modes indexed by \(k\in\mathbb Z^2\), and each modal coefficient \(\hat u(k,t)\) evolves independently according to a finite-dimensional linear system \(\partial_t\hat u(k,t)=A(k)\hat u(k,t)\), where \(A(k)\in\mathbb C^{d_u\times d_u}\) is the matrix induced by \(\mathcal L_x\) at frequency \(k\). The finite-time propagation is \(\hat u(k,t+\Delta t)=\exp(\Delta t A(k))\hat u(k,t)\). This decomposition reveals a crucial structural bias: for linear systems, the natural propagation space is the spatial Fourier basis, where temporal evolution is decoupled into independent, low-dimensional subsystems~\cite{dft_pde_linear,trefethen2000spectral,canuto2006spectral}. While this exact decoupling holds only in the linear translation-invariant setting, it provides a useful template for organizing latent propagation in more general PDEs. It suggests that an appropriate spatial basis can expose an explicit and structured propagation law for PDE evolution.


\noindent\textbf{Nonlinear coupling and finite-band closure.}
For general PDEs governed by Eq.~\eqref{eq:general_pde}, nonlinear terms induce interactions across Fourier modes and break the strict mode-wise independence of the linear case. For example, the Fourier coefficient of a quadratic term \(u^2\) at mode \(k\) is given by
\(\widehat{u^2}(k)=\sum_{p+q=k}\hat u(p)\hat u(q)\), where \(p\) and \(q\) denote interacting modes. Thus, the evolution of a mode is generally affected by combinations of other modes, allowing energy transfer across spectral bands. Since practical propagators retain only a finite spectral band for tractability and stability, the retained dynamics are non-closed: unresolved modes can still affect the retained band through nonlinear coupling. This calls for a closure term that compensates for energy transfer and unresolved interactions between retained and truncated modes~\cite{pde_refiner,sagaut2006large,pope2001turbulent,chorin1998optimal}. Rather than reconstructing discarded modes, the closure refines retained spectral evolution and reduces spectral distortion during long rollout.


Synthesizing these insights, we model \(\mathcal G\) through three coupled objectives: (i) constructing a time-consistent spatial representation space shared across all time slices; (ii) projecting this rich spatial space to a compact propagation state to reduce the conflict between reconstruction fidelity and rollout regularity; (iii) structuring the latent propagator \(\Phi\) in this compact state explicitly as a modal evolution rule, with a frequency-conditioned linear backbone for retained modes and a nonlinear closure for coupled and truncated interactions. 

Consequently, we factorize the learned operator as
 $\widehat{\mathcal G}
:= \mathcal{D} \circ \mathcal{R} \circ \Phi \circ \mathcal{P} \circ \mathcal{E}$, where \(\mathcal E\), \(\mathcal P\), \(\Phi\), \(\mathcal R\), and \(\mathcal D\) instantiate spatial-basis analysis, projection to a compact propagation state, structured spectral propagation, inverse projection, and physical-field synthesis, respectively.

\section{Structured Spectral Propagator Learning}
\noindent\textbf{Problem formulation.} 
Let \(u_n:=u(\cdot,t_n)\in\mathcal U\) denote the solution state at time \(t_n\). For a fixed sequence length \(T\), we denote the state sequence starting at \(t_n\) by
\(\mathbf u_n := (u_n,\ldots,u_{n+T-1})\).
 The PDE evolution induces a finite-time propagator \(\mathcal G_T\), with \(\mathbf u_{n+t}=\mathcal{G_T}(\mathbf u_n)\). We approximate this propagator via five maps: \(\widehat{\mathcal G}_T:=\mathcal D\circ\mathcal R\circ\Phi\circ\mathcal P\circ\mathcal E\). 



\paragraph{Encoder.}
The encoder \(\mathcal E\) constructs a shared spatial representation from physical solution states while preserving fine-grained spatial details needed for later synthesis. Given a temporal state sequence \(\mathbf u_n\), the encoder is applied to each time slice, yielding \(\mathbf h_n=(h_n,\ldots,h_{n+T-1})\), where \(h_j=\mathcal E(u_j)\in\mathcal H \subset\mathbb R^{C_s\times n_x\times n_y}\). This frame-wise design prevents the representation space \(\mathcal H\) from entangling spatial features with short-range temporal advancement. As a result, \(\mathcal H\) serves as a spatial observation space, leaving temporal evolution modeled explicitly by the structured propagator \(\Phi\).

In implementation, each grid-based state \(u_j\in\mathbb R^{d_u\times N_x\times N_y}\) is concatenated with a normalized coordinate grid \(\gamma\in\mathbb R^{2\times N_x\times N_y}\), yielding \(\hat u_j\in\mathbb R^{(d_u+2)\times N_x\times N_y}\). A local convolutional lift maps \(\hat u_j\) to \(C_s\) feature channels and optionally reduces the resolution to \(n_x\times n_y\), extracting local geometry and cross-variable correlations from the physical field. Cascaded factorized spectral blocks then apply one-dimensional Fourier transforms along each spatial axis, and mix retained modes across channels, providing lightweight spectral calibration. A final channel attention module reweights feature channels according to their global relevance, yielding \(h_j=\mathcal E(u_j)\in\mathcal H\). Thus, the encoder provides a rich and frequency-aware spatial representation for subsequent projection and structured propagation.

\paragraph{Projector.}
The projector \(\mathcal P\) maps the rich spatial representation \(h_j\in\mathcal H\) to a compact propagation state \(z_j\in\mathcal Z\). Its role is to separate the spatial details for synthesis from the dynamically focused state used for recurrent dynamics, mitigating the conflict between reconstruction fidelity and propagation regularity. In implementation, \(\mathcal P\) is a pointwise channel projection implemented by a \(1\times1\) convolution. Given \(h_j\in\mathbb R^{C_s\times n_x\times n_y}\), \(\mathcal P\) maps it to \(z_j=\mathcal P(h_j)\in\mathbb R^{C_z\times n_x\times n_y}\), where \(C_z<C_s\). Thus, $\mathcal{P}$ preserves the latent spatial resolution while reducing the channel dimension from \(C_s\) to \(C_z\). This channel bottleneck suppresses redundant representation degrees of freedom and encourages the propagated state to retain dynamically relevant features. Applied frame-wise to \(\mathbf h_n=(h_n,\ldots,h_{n+T-1})\), it yields the compact state sequence \(\mathbf z_n=(z_n,\ldots,z_{n+T-1})\) on which the structured propagator operates. Preserving the latent spatial grid is important for the subsequent Fourier-domain modal update and avoids discarding spatial phase information before rollout.
\paragraph{Propagator.}
The propagator \(\Phi\) advances the compact sequence \(\mathbf z_n=(z_n,\ldots,z_{n+T-1})\) to its future sequence \(\tilde{\mathbf z}_{n+T}=\Phi(\mathbf z_n)\) in the propagation space \(\mathcal Z\), where \(z_j\in\mathbb R^{C_z\times n_x\times n_y}\). To impose a spectral organization on the update, we apply a spatial Fourier transform \(\mathcal F_s\) to each state and restrict the coefficients to a retained low-frequency set \(\Lambda=\Lambda_x\times\Lambda_y\), where \(|\Lambda_x|=m_x\) and \(|\Lambda_y|=m_y\). With \(\Pi_\Lambda\) denoting Fourier truncation, we obtain \(\widehat{\mathbf z}_{n,\Lambda}:=\Pi_\Lambda\mathcal F_s(\mathbf z_n)\in\mathbb C^{T\times C_z\times m_x\times m_y}\). For each retained mode \(k\in\Lambda\) and channel \(\ell\), the temporal modal vector \(q^{(k,\ell)}\in\mathbb C^T\) collects the Fourier coefficients of this mode across the \(T\) time slices. The retained spectral block is then updated by a frequency-conditioned linear backbone \(K_\theta\) and a nonlinear spectral closure \(G_\psi\).

\emph{Frequency-conditioned linear backbone.}
The linear backbone \(K_\theta\) models the dominant temporal evolution of retained spectral modes. Motivated by the modal organization of spectral PDE evolution mentioned above, we parameterize the latent update as mode-wise linear temporal dynamics whose action is adaptively modulated by frequency. For each latent channel \(\ell\), a single temporal evolution matrix \(\bar K_\ell\in\mathbb C^{T\times T}\) is shared across all retained modes, while mode-specific behavior is introduced through a bounded frequency-dependent gate, $
K_\theta^{(k,\ell)}q
=
M_\theta(k,\ell)\bar K_\ell q$, where
$M_\theta(k,\ell)
=
1+\beta\tanh(a_\theta(\xi(k))_\ell)$. Here, \(\xi(k)=(k_x,k_y,\|k\|,\cos\theta_k, \sin\theta_k)\), \(\theta_k=\operatorname{atan2}(k_y,k_x)\), and \(a_\theta\) is a shared MLP. The feature \(\xi(k)\) encodes both the Cartesian position and polar geometry of the spatial frequency, allowing the same temporal backbone to adapt across wavelengths and orientations. This yields a parameter-efficient backbone design: instead of learning a separate \(T\times T\) matrix for every frequency-channel pair, it uses only \(C_z\) temporal matrices together with a shared gating network. The bounded form \(1+\beta\tanh(\cdot)\) limits unconstrained frequency-dependent deviations during recurrent rollout. 

To regularize the temporal backbone during rollout, we impose the normality penalty
\(\mathcal L_{\mathrm{norm}}=\frac{1}{C_z}\sum_{\ell=1}^{C_z}
\|\bar K_\ell\bar K_\ell^{H}-\bar K_\ell^{H}\bar K_\ell\|_F^2\).
This penalty encourages each temporal evolution matrix \(\bar K_\ell\) to approach a normal matrix, for which repeated action is controlled by eigenvalues through unitary diagonalization. By discouraging transient amplification induced by non-orthogonal eigenvectors, it improves the conditioning of autoregressive temporal updates.

\emph{Nonlinear spectral closure.}
The closure \(G_{\psi}\) compensates for residual effects not represented by the linear modal backbone, including unresolved nonlinear coupling and energy transfer among retained modes, and finite-band truncation effects. It operates on the retained spectral block \(\widehat{\mathbf z}_{n,\Lambda}\). In implementation, we split by the complex coefficients into real and imaginary parts, concatenate time and channel axes, and apply a shallow convolutional network \(\mathcal C_\psi\) over the retained frequency plane \((m_x,m_y)\).
The convolutional structure aggregates nearby retained modes, while the merged temporal-channel dimension provides joint time and channel context. We initialize \(G_\psi\) near zero so that early training is dominated by the linear backbone. To encourage complementary updates, we penalize the alignment between the linear and closure directions by $\mathcal L_{\mathrm{orth}}
=
\frac{|\langle \Delta_K,\Delta_G\rangle|^2}
{(\|\Delta_K\|^2+\epsilon)(\|\Delta_G\|^2+\epsilon)}$, where \(\Delta_K=K_\theta(\widehat{\mathbf z}_{n,\Lambda})-\widehat{\mathbf z}_{n,\Lambda}\) denotes the update direction induced by the linear backbone, and \(\Delta_G=G_\psi(\widehat{\mathbf z}_{n,\Lambda})\) denotes the nonlinear closure correction. This discourages $G_\psi$ from duplicating the linear update and encourages it to focus on residual nonlinear and truncation-induced effects.


\emph{Residual spectral update.}
Rather than using a single unconstrained transition on the retained spectral block, we formulate propagation as the discretization of a learned latent flow in the retained spectral space~\cite{neuralode}. Starting from \(\widehat{\mathbf z}_{n,\Lambda}^{(0)}=\widehat{\mathbf z}_{n,\Lambda}\), we perform \(N_{\mathrm{sub}}\) residual substeps:
\[
\widehat{\mathbf z}_{n,\Lambda}^{(s+1)}
=
\widehat{\mathbf z}_{n,\Lambda}^{(s)}
+
\Delta\tau
\left[
\alpha\bigl(K_\theta(\widehat{\mathbf z}_{n,\Lambda}^{(s)})-\widehat{\mathbf z}_{n,\Lambda}^{(s)}\bigr)
+
\lambda_g G_\psi(\widehat{\mathbf z}_{n,\Lambda}^{(s)})
\right],
\qquad
s=0,\ldots,N_{\mathrm{sub}}-1 .
\]
This update can be viewed as an explicit Euler discretization of the latent-time dynamics \(\partial_\tau \widehat{\mathbf z}_{\Lambda}=\alpha(K_\theta(\widehat{\mathbf z}_{\Lambda})-\widehat{\mathbf z}_{\Lambda})+\lambda_gG_\psi(\widehat{\mathbf z}_{\Lambda})\), where \(\tau\) is an internal evolution variable in the latent spectral space. Accordingly, \(\Delta\tau\) is the latent integration step size, while \(N_{\mathrm{sub}}\Delta\tau\) controls the total internal propagation depth. The backbone term provides the dominant mode-wise linear increment, and the closure supplies a state-dependent nonlinear correction. In our experiments, \(\alpha\) and \(\lambda_g\) are fixed scalar weights, set to \(\alpha=1\) and \(\lambda_g=1\). After the final substep, the updated retained block is inserted back into the spectral representation and transformed to \(\mathcal Z\) by inverse Fourier transform, yielding \(\tilde{\mathbf z}_{n+T}=\Phi(\mathbf z_n)\).


\paragraph{Inverse Projector.}
The inverse projector \(\mathcal R\) lifts the propagated compact state back to the spatial representation space before physical-field synthesis. Given the propagated compact sequence \(\tilde{\mathbf z}_{n+T}=(\tilde z_{n+T},\ldots,\tilde z_{n+2T-1})\), the reconstructor is applied to each state as \(\tilde h_j=\mathcal R(\tilde z_j)\), yielding \(\tilde{\mathbf h}_{n+T}=(\tilde h_{n+T},\ldots,\tilde h_{n+2T-1})\). In implementation, \(\mathcal R\) is a \(1\times1\) convolution that maps \(\tilde z_j\in\mathbb R^{C_z\times n_x\times n_y}\) to \(\tilde h_j\in\mathbb R^{C_s\times n_x\times n_y}\). Thus, \(\mathcal R\) restores the channel capacity required by the decoder while preserving the latent spatial resolution.

\paragraph{Decoder.}
The decoder \(\mathcal D\) maps the lifted representation \(\tilde h_j\in\mathcal H\) back to the physical solution variables. Applied frame-wise to \(\tilde{\mathbf h}_{n+T}\), it yields the predicted sequence \(\tilde{\mathbf u}_{n+T}=(\tilde u_{n+T},\ldots,\tilde u_{n+2T-1})\), where \(\tilde u_j=\mathcal D(\tilde h_j)\). In implementation, \(\mathcal D\) combines channel attention, a factorized spectral branch, a local convolutional branch, and a readout head. Channel attention first recalibrates the lifted feature channels. The spectral branch synthesizes global structure, while the local branch refines spatial details and residual high-frequency components not explicitly evolved in the compact spectral state. The readout head maps the refined features to \(d_u\) physical channels, yielding \(\tilde u_j\in\mathbb R^{d_u\times N_x\times N_y}\).

\paragraph{Training objective.}
The model is trained by autoregressive rollout on temporal state sequences. Let \(\mathbf u^{(r)}:=\mathbf u_{n+rT}\) denote the ground-truth sequence at rollout step \(r\), and let \(N_{\mathrm{roll}}\) be the number of rollout steps used for training. Starting from \(\tilde{\mathbf u}^{(0)}=\mathbf u^{(0)}\), the model recursively computes \(\mathbf z_{\mathrm{in}}^{(r)}=\mathcal P(\mathcal E(\tilde{\mathbf u}^{(r)}))\), \(\tilde{\mathbf z}^{(r+1)}=\Phi(\mathbf z_{\mathrm{in}}^{(r)})\), and \(\tilde{\mathbf u}^{(r+1)}=\mathcal D(\mathcal R(\tilde{\mathbf z}^{(r+1)}))\), for \(r=0,\ldots,N_{\mathrm{roll}}-1\). For latent supervision, we define the encoded target compact sequence as \(\mathbf z_{\star}^{(r+1)}=\mathcal P(\mathcal E(\mathbf u^{(r+1)}))\). The objective combines reconstruction, latent rollout, and physical rollout losses. This hierarchical supervision encourages a propagation-compatible representation and constrains rollout errors in both the compact latent space and the physical solution space. The training objective is
\[
\mathcal L
=
\lambda_{\mathrm{rec}}\mathcal L_{\mathrm{rec}}
+
\frac{1}{N_{\mathrm{roll}}}
\sum_{r=0}^{N_{\mathrm{roll}}-1}
\left(
\lambda_{\mathrm{lat}}\mathcal L_{\mathrm{lat}}^{(r)}
+
\lambda_{\mathrm{phy}}\mathcal L_{\mathrm{phy}}^{(r)}\right)
+
\lambda_{\mathrm{norm}}\mathcal L_{\mathrm{norm}}
+
\lambda_{\mathrm{orth}}\mathcal L_{\mathrm{orth}} .
\]
The reconstruction loss \(\mathcal L_{\mathrm{rec}}=\|\mathbf u^{(0)}-\mathcal D(\mathcal E(\mathbf u^{(0)}))\|_2^2\) preserves spatial information through the encoder--decoder path without temporal propagation. The latent rollout loss \(\mathcal L_{\mathrm{lat}}^{(r)}=\|\tilde{\mathbf z}^{(r+1)}-\mathbf z_{\star}^{(r+1)}\|_2^2\) aligns the propagated compact sequence with the encoded target sequence. The physical rollout loss \(\mathcal L_{\mathrm{phy}}^{(r)}=\|\tilde{\mathbf u}^{(r+1)}-\mathbf u^{(r+1)}\|_2^2\) constrains prediction error in the solution space. The regularizers \(\mathcal L_{\mathrm{norm}}\) and \(\mathcal L_{\mathrm{orth}}\) are defined in the propagator, controlling the normality of the temporal backbone and the alignment between the backbone and closure increments, respectively. We set fixed weights \(\lambda_{\mathrm{rec}}=1\), \(\lambda_{\mathrm{lat}}=1\), \(\lambda_{\mathrm{phy}}=1\), and \(\lambda_{\mathrm{norm}}=\lambda_{\mathrm{orth}}=0.01\) in all experiments.

\section{Experiments}

We run experiments across different dynamical regimes to assess whether the proposed framework improves stability and spectral consistency under recursive rollout.
(i) \textbf{Long-horizon forecasting accuracy.} We evaluate autoregressive prediction on Shallow--Water (SW), Reaction--Diffusion (RD), and Navier--Stokes (NS) benchmarks~\cite{takamoto2022pdebench,FNO}, comparing against representative spectral and latent neural operator baselines in terms of trajectory accuracy, worst-case local error, and spectral fidelity. 
(ii) \textbf{Temporal extrapolation.} We test whether the learned propagator remains reliable beyond the supervised training horizon,  indicating a reusable evolution rule rather than fit to the observed rollout range. 
(iii) \textbf{Ablation studies.} We assess the contribution of the linear modal backbone \(K_\theta\), nonlinear closure \(G_\psi\), compact propagation space induced by \(\mathcal P,\mathcal R\), frame-wise encoding, and spectral encoder--decoder design. For SSP, the results are reported as the mean over three independent random seeds, with standard deviations provided in Appendix. 

\begin{table}[t]
  \centering
  \caption{Quantitative comparison on long-horizon PDE benchmarks. Best results are in \textbf{bold}, and second-best results are \underline{underlined}. - indicates that the corresponding model failed to train under this setting.}
  \label{tab:long-horizon-prediction-table}

  \footnotesize
  \setlength{\tabcolsep}{3.5pt}
  \renewcommand{\arraystretch}{1.08}

  \begin{tabular}{@{}lccccccccc@{}}
    \toprule
    \multirow{2}{*}{\textbf{Method}}
    & \multicolumn{3}{c}{\textbf{Navier--Stokes}}
    & \multicolumn{3}{c}{\textbf{Shallow--Water}}
    & \multicolumn{3}{c}{\textbf{Reaction--Diffusion}} \\
    \cmidrule(lr){2-4} \cmidrule(lr){5-7} \cmidrule(lr){8-10}
    & \textbf{L2} & \textbf{$E_{\max}$} & \textbf{$f_{\mathrm{low}}$}
    & \textbf{L2} & \textbf{$E_{\max}$} & \textbf{$f_{\mathrm{low}}$}
    & \textbf{L2} & \textbf{$E_{\max}$} & \textbf{$f_{\mathrm{low}}$} \\
    \midrule
    FNO
    & 1.42E-1 & 6.68E-1 & 5.94E-2
    & 5.10E-3 & 8.00E-2 & 3.37E-4
    & 2.57E-1 & 6.72E-2 & 1.28E-2 \\

    U-FNO
    & 1.50E-1 & 1.44E+0 & 2.38E-2
    & 4.99E-3 & 6.04E-2 & 1.36E-3
    & \underline{1.46E-2} & 3.53E-2 & \underline{1.88E-4} \\

    F-FNO
    & \underline{7.57E-2} & 6.87E-1 & \underline{1.09E-2}
    & \underline{9.28E-4} & \underline{1.69E-2} & 6.99E-5
    & 1.52E-2 & \underline{2.02E-2} & 2.69E-4 \\

    \midrule
    UNO
    & 1.36E-1 & 1.19E+0 & 2.16E-2
    & 1.76E-3 & 1.71E-2 & 7.40E-5
    & 8.38E-2 & 3.29E-1 & 4.18E-3 \\

    KNO
    & 7.77E-2 & \textbf{3.82E-1} & 5.60E-2
    & 9.76E-3 & 3.12E-2 & 3.89E-3
    & 2.51E-1 & 6.85E-2 & 1.09E-2 \\

    LNO
    & 1.32E-1 & 1.16E+0 & 2.18E-2
    & 2.31E-3 & 2.11E-2 & 6.12E-5
    & - & - & - \\

    CALM
    & 8.82E-2 & 9.96E-1 & 1.40E-2
    & 2.25E-3 & 2.41E-2 & \underline{3.61E-5}
    & 1.72E-1 & 2.30E-1 & 5.40E-4 \\

    Ours
    & \textbf{5.50E-2} & \underline{5.38E-1} & \textbf{5.49E-3}
    & \textbf{6.56E-4} & \textbf{9.72E-3} & \textbf{2.53E-5}
    & \textbf{7.46E-3} & \textbf{6.71E-3} & \textbf{1.48E-4} \\

    \bottomrule
  \end{tabular}
\end{table}

\begin{figure}[t]
		\centering
		\includegraphics[width=.99\linewidth]{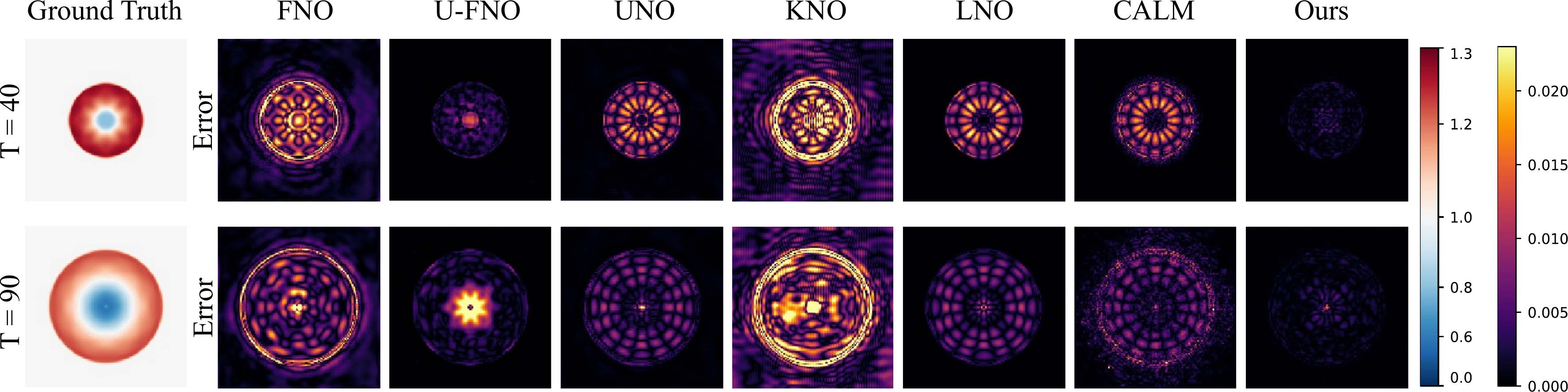}
		\caption{Ground-truth states and absolute-error maps on the Shallow--Water benchmark at rollout steps \(t=40\) and \(t=90\). Leftmost column: ground truth. Remaining columns: pointwise absolute-error maps for different methods.}
		\label{error_sw}
\end{figure}

\subsection{Long-horizon forecasting accuracy}

We first evaluate long-horizon forecasting accuracy to assess whether a learned propagator remains stable under repeated composition, rather than only fitting short local transitions. We consider the three PDE benchmarks introduced above, covering different dynamical characteristics. We refer to Appendix for dataset details. For SW and RD, each model observes 10 input frames and predicts the next 90 frames autoregressively. For NS, each model observes 4 input frames and predicts the following 12 frames. We report time-averaged metrics that capture complementary aspects of rollout quality: relative \(L_2\) error for global trajectory accuracy, \(E_{\max}\) for worst-case pointwise deviation, and \(f_{\mathrm{low}}\) for preservation of dominant low-frequency spectral modes.

Table~\ref{tab:long-horizon-prediction-table} compares our method with representative spectral and latent neural operator baselines. On SW and RD, our method achieves the best results across all reported metrics, showing consistent gains in global accuracy, worst-case local error, and low-frequency spectral fidelity. On NS, it obtains the lowest \(L_2\) and \(f_{\mathrm{low}}\), while KNO gives a lower \(E_{\max}\); however, KNO has substantially larger global and spectral errors, indicating weaker rollout consistency. The full spectral evaluation in Appendix further shows that our method also achieves the best mid- and high-frequency errors across all three benchmarks, suggesting that the structured propagator preserves dominant dynamics without sacrificing finer spectral components. 


Figure~\ref{error_sw} presents ground-truth states and absolute-error maps on SW at rollout steps \(t=40\) and \(t=90\), corresponding to intermediate and late autoregressive prediction stages. This benchmark requires the model to preserve the radial envelope, moving wavefront, and global symmetry as errors accumulate over time. At \(t=40\), several baselines already exhibit structured residual patterns, including ringing artifacts, lattice-like errors, and deviations concentrated near the wavefront, whereas our method maintains weaker and more localized errors. By \(t=90\), these artifacts become more pronounced for the baselines, indicating progressive degradation of wavefront location and radial morphology under recursive rollout. In contrast, our method produces the smallest-magnitude and least structured error field, suggesting better control of error accumulation and spatial error organization. These visual observations are consistent with Table~\ref{tab:long-horizon-prediction-table}. Additional error visualizations on the remaining benchmarks are provided in the Appendix.

\begin{table}[t]
  \centering
  \caption{Temporal extrapolation results on the Shallow--Water benchmark. Errors are computed only over rollout steps 51--90, which lie beyond the supervised training horizon. Best results are in \textbf{bold}, and second-best results are \underline{underlined}.}
  \label{tab:sw_extrapolation}

  \footnotesize
  \setlength{\tabcolsep}{3.4pt}
  \renewcommand{\arraystretch}{1.08}

  \begin{tabular}{@{}lccccccc@{}}
    \toprule
    \textbf{Method}
    & \textbf{L2}
    & \textbf{$E_{\max}$}
    & \textbf{BRMS}
    & \textbf{$f_{\mathrm{low}}$}
    & \textbf{$f_{\mathrm{mid}}$}
    & \textbf{$f_{\mathrm{high}}$}
    & \textbf{$f_{\mathrm{mse}}$} \\
    \midrule
    FNO
    & 1.83E+02 & 3.29E+02 & 4.86E+01
    & 1.16E+01 & 7.30E+00 & 6.63E-01 & 2.17E+00 \\

    U-FNO
    & 1.28E+00 & 1.02E+01 & 1.37E-02
    & 2.40E-01 & 8.51E-02 & 8.07E-03 & 3.22E-02 \\

    F-FNO
    & \textbf{4.36E-02} & 4.69E-01 & \underline{6.51E-03}
    & \textbf{7.71E-03} & \underline{5.04E-03} & 1.68E-03 & \underline{1.68E-03} \\

    \midrule
    UNO
    & 3.56E+11 & 4.21E+11 & 1.47E+11
    & 6.01E+10 & 9.50E+09 & 1.26E+09 & 5.97E+09 \\

    KNO
    & 8.19E-02 & 9.96E-01 & 1.90E-02
    & 4.71E-02 & 1.94E-02 & 8.99E-03 & 2.72E-02 \\

    LNO
    & 1.71E+00 & 2.96E+02 & 1.13E+01
    & 2.43E+00 & 1.05E+00 & 1.18E+00 & 1.24E+00 \\

    CALM
    & 6.41E-02 & \underline{3.68E-01} & 2.17E-02
    & 1.56E-02 & 5.36E-03 & \underline{7.98E-04} & 2.29E-03 \\

    Ours
    & \underline{4.49E-02} & \textbf{2.55E-01} & \textbf{5.46E-04}
    & \underline{1.24E-02} & \textbf{2.02E-03} & \textbf{2.59E-04} & \textbf{1.24E-03} \\
    \bottomrule
  \end{tabular}
\end{table}

\begin{figure}[t]
		\centering
		\includegraphics[width=.99\linewidth]{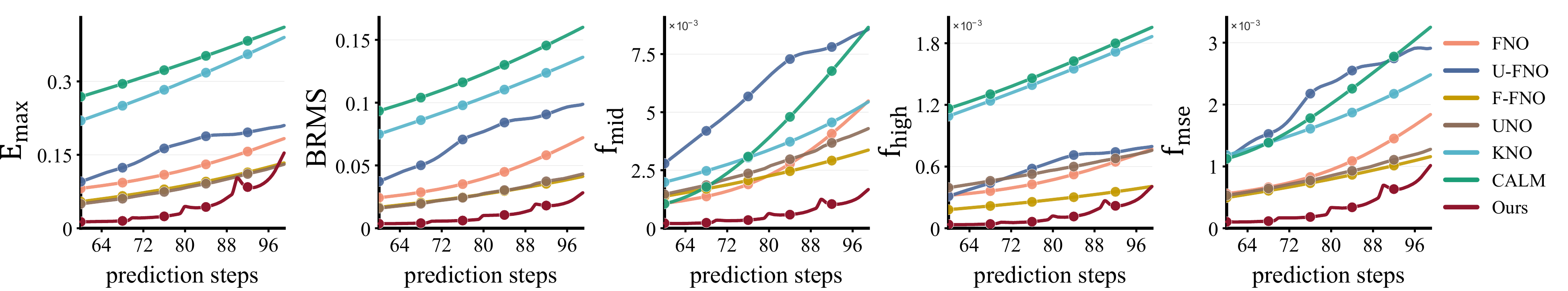}
        \caption{Temporal extrapolation error curves on the Reaction--Diffusion benchmark over rollout steps 51--90. Each curve reports the mean error at each extrapolation step; lower curves indicate slower error accumulation beyond the supervised training horizon.}
		\label{extra_rdf}
\end{figure}

\subsection{Temporal extrapolation beyond the training horizon}

We next evaluate whether the learned propagator remains reliable beyond the supervised rollout range on unseen test trajectories. All models are trained on training trajectories with 10 input frames and 50 supervised prediction steps. At test time, each model receives 10 input frames from the test set, corresponding to unseen initial conditions, and is autoregressively rolled out for 90 steps. Errors are reported only on steps 51--90, which lie beyond the supervised training horizon. This setting probes whether the model has learned a reusable evolution law that generalizes across both initial conditions and temporal ranges. We evaluate extrapolation using complementary physical-domain metrics and spectral-domain metrics. Relative \(L_2\), \(E_{\max}\), and BRMS measure global trajectory error, worst-case local deviation, and boundary consistency under autoregressive
rollout, respectively, while \(f_{\mathrm{low}}\), \(f_{\mathrm{mid}}\), \(f_{\mathrm{high}}\), and \(f_{\mathrm{mse}}\) quantify frequency-domain distortion.


Table~\ref{tab:sw_extrapolation} reports temporal extrapolation results on SW, a substantially more demanding regime than supervised rollout. UNO, FNO and LNO show severe error growth, indicating that their learned updates do not remain bounded beyond the training horizon. F-FNO achieves the best \(f_{\mathrm{low}}\), showing strong dominant-mode alignment, but its larger BRMS and high-frequency error suggest weaker boundary
consistency and more pronounced fine-scale spectral distortion. CALM remains stable with competitive high-frequency error, but is weaker in trajectory and mid-frequency fidelity. In contrast, our method achieves the best \(E_{\max}\), BRMS, \(f_{\mathrm{mid}}\), \(f_{\mathrm{high}}\), and \(f_{\mathrm{mse}}\), with substantially lower BRMS than all baselines. This suggests SSP is particularly effective at controlling local error growth, maintaining boundary consistency, and preserving finer spectral components.

Figure~\ref{extra_rdf} further shows step-wise extrapolation errors on RD over the same beyond-horizon interval. Across \(E_{\max}\), BRMS, \(f_{\mathrm{mid}}\), \(f_{\mathrm{high}}\), and \(f_{\mathrm{mse}}\), our method maintains the lowest error curves and the consistently lower error growth throughout the rollout. The step-wise curves confirm that the improvement persists throughout the extrapolation interval, rather than being an artifact of averaging. Full numerical metrics in both physical and spectral domains are reported in Appendix. 

\begin{table}[t]
  \centering
  \caption{Ablation study on Navier--Stokes, Shallow--Water, and
  Reaction--Diffusion. w/o denotes removing the corresponding component from the full model. Best results are in \textbf{bold}, and second-best results are
  \underline{underlined}.}
  \label{tab:three_ablation}

  \footnotesize
  \setlength{\tabcolsep}{3.0pt}
  \renewcommand{\arraystretch}{1.08}

  \begin{tabular}{@{}lccccccccc@{}}
    \toprule
    \multirow{2}{*}{\textbf{Method}}
    & \multicolumn{3}{c}{\textbf{Navier--Stokes}}
    & \multicolumn{3}{c}{\textbf{Shallow--Water}}
    & \multicolumn{3}{c}{\textbf{Reaction--Diffusion}} \\
    \cmidrule(lr){2-4} \cmidrule(lr){5-7} \cmidrule(lr){8-10}
    & \textbf{L2} & \textbf{\(E_{\max}\)} & \textbf{\(f_{\mathrm{low}}\)}
    & \textbf{L2} & \textbf{\(E_{\max}\)} & \textbf{\(f_{\mathrm{low}}\)}
    & \textbf{L2} & \textbf{\(E_{\max}\)} & \textbf{\(f_{\mathrm{low}}\)} \\
    \midrule
    Full
    & \textbf{5.50E-2} & \textbf{5.38E-1} & \textbf{5.49E-3}
    & \textbf{6.56E-4} & \textbf{9.72E-3} & \textbf{2.53E-5}
    & \textbf{7.46E-3} & \textbf{6.71E-3} & \textbf{1.48E-4} \\

    w/o \(K_\theta\)
    & 6.20E-2 & 6.48E-1 & 6.41E-3
    & \underline{9.09E-4} & \underline{1.30E-2} & 8.50E-5
    & 1.43E-2 & 1.15E-2 & 4.80E-4 \\

    w/o \(G_\psi\)
    & \underline{6.04E-2} & \underline{5.93E-1} & \underline{5.92E-3}
    & 1.78E-3 & 2.16E-2 & 7.70E-5
    & \underline{1.12E-2} & \underline{8.12E-3} & \underline{2.90E-4} \\

    w/o \(\mathcal{P},\mathcal{R}\)
    & 6.07E-2 & 6.21E-1 & 6.11E-3
    & 1.05E-3 & 1.38E-2 & \underline{4.90E-5}
    & 2.41E-2 & 2.20E-2 & 1.40E-3 \\

    \(T{\to}C\) Enc
    & 8.35E-2 & 9.43E-1 & 8.18E-3
    & 2.06E-3 & 3.36E-2 & 1.01E-4
    & 6.41E-2 & 3.99E-2 & 2.24E-3 \\

    Conv Enc-Dec
    & 1.53E-1 & 1.48E+0 & 2.57E-2
    & 1.08E-3 & 1.76E-2 & 5.50E-5
    & 1.95E-2 & 1.68E-2 & 8.50E-4 \\
    \bottomrule
  \end{tabular}
\end{table}

\subsection{Ablation studies}

We conduct ablation studies on NS, SW, and RD to examine the roles of the
analysis--propagation--synthesis components. We remove the linear backbone
\(K_\theta\), the nonlinear closure \(G_\psi\), and the projector pair
\(\mathcal P,\mathcal R\). We further test two architectural variants: \(T{\to}C\) Enc, which treats the
input history as a channel-stacked field before encoding, instead of using
frame-wise spatial encoding, and Conv Enc-Dec, which replaces the
spectral-factorized encoder--decoder with convolutional modules under the same
resolution schedule. Table~\ref{tab:three_ablation} shows that the full model achieves the best
results across all reported metrics. Removing \(K_\theta\) or \(G_\psi\)
degrades both physical and spectral accuracy, supporting their complementary
roles as the linear modal backbone and nonlinear closure. Removing
\(\mathcal P,\mathcal R\) also hurts performance, indicating the benefit of a
separate compact propagation space. The \(T{\to}C\) Enc variant performs
substantially worse under the same propagation pipeline, suggesting that
temporal-to-channel stacking is less compatible with SSP than frame-wise
analysis followed by explicit propagation in \(\Phi\). Conv Enc-Dec remains competitive after replacing the spectral encoder--decoder,
indicating that the advantage of SSP is not tied to a specific analysis--synthesis
backbone but can extend to alternative encoder--decoder designs.

\section{Conclusion}
This work proposes SSP for long-horizon PDE forecasting, an analysis--propagation--synthesis framework that separates time-consistent spatial representation, compact propagation, and physical-field synthesis. SSP structures latent spectral evolution with a frequency-conditioned linear backbone and a nonlinear closure over retained modes. Experiments on three PDE benchmarks demonstrate improved long-horizon accuracy, spectral preservation, and beyond-horizon stability, supporting structured latent spectral propagation as an effective principle for long-term physical field prediction.

\bibliographystyle{ieeetr}
\bibliography{main}

\end{document}